\documentclass[11pt]{article}

\usepackage[a4paper,margin=1in]{geometry}
\usepackage[utf8]{inputenc}
\usepackage[T1]{fontenc}
\usepackage{setspace}
\usepackage{parskip}
\usepackage{graphicx}
\usepackage{xcolor}
\usepackage{titlesec}
\usepackage{booktabs}
\usepackage{multirow}
\usepackage{fancyhdr}
\usepackage{natbib}
\usepackage{amsmath,amsfonts,amsthm,amssymb}
\usepackage{hyperref}
\usepackage{cleveref}
\usepackage{lmodern}
\usepackage{tcolorbox}
\usepackage{csquotes}

\definecolor{sfblue}{HTML}{00A1E0}      
\definecolor{sfnavy}{HTML}{032D60}      
\definecolor{sfgray}{HTML}{706E6B}      
\definecolor{sflightblue}{HTML}{EAF5FC} 

\hypersetup{
  colorlinks=true,
  linkcolor=sfblue,
  citecolor=sfnavy,
  urlcolor=sfblue
}

\renewcommand{\headrulewidth}{0.4pt} 
\renewcommand{\footrulewidth}{0pt} 

\fancypagestyle{plain}{
  \fancyhf{} 
  \fancyfoot[C]{\textcolor{sfgray}{\thepage}} 
  \renewcommand{\headrulewidth}{0pt}
}

\usepackage{xspace}

\newcommand{\github}{\raisebox{-1.5pt}{\includegraphics[height=1.05em]{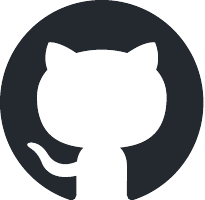}}\xspace}

\makeatletter
\renewcommand{\maketitle}{
  \thispagestyle{plain} 
  \noindent
  \vspace*{-10pt}
  \noindent\raisebox{0pt}[0pt][0pt]{\includegraphics[height=1cm]{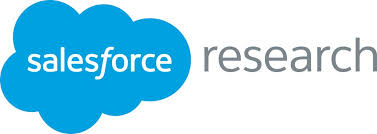}}

  \vspace{-5pt}
  \color{sfgray}\rule{\linewidth}{0.6pt}

  \vspace{10pt}
  {\Huge\bfseries\color{sfnavy} \@title \par}
  \vspace{0.5em}
  {\large \@author \par}
  \vspace{0.2em}
  {\normalsize $\dagger$Core Contributor\quad | \quad Salesforce AI Research \quad | \quad \@date \par}
  \vspace{1.2em}
}
\makeatother

\titleformat{\section}{\large\bfseries\color{sfnavy}}{\thesection}{1em}{}
\titleformat{\subsection}{\normalsize\bfseries\color{sfnavy}}{\thesubsection}{1em}{}

\title{Prompt Optimization Via Diffusion Language Models}
\author{Shiyu Wang$\dagger$, Haolin Chen$\dagger$, Liangwei Yang, Jielin Qiu, Rithesh Murthy, Ming Zhu, Zixiang Chen, Silvio Savarese, Caiming Xiong, Shelby Heinecke, Huan Wang}
\date{\today}

\usepackage{amsmath,amsfonts,bm}

\def\eqref#1{equation~\ref{#1}}

\def\1{\bm{1}}

\DeclareMathAlphabet{\mathsfit}{\encodingdefault}{\sfdefault}{m}{sl}
\SetMathAlphabet{\mathsfit}{bold}{\encodingdefault}{\sfdefault}{bx}{n}

\begin{document}

\maketitle

\begin{tcolorbox}[colback=sflightblue!60,
                  colframe=sfblue,
                  boxrule=0.6pt,
                  arc=2mm,
                  left=6pt,right=6pt,top=6pt,bottom=6pt]
\textbf{Abstract.} 
We propose a diffusion-based framework for prompt optimization that leverages Diffusion Language Models (DLMs) to iteratively refine system prompts through masked denoising. By conditioning on interaction traces, including user queries, model responses, and optional feedback, our method enables flexible, span-level prompt updates without requiring gradient access or modifying the downstream language model. Across diverse benchmarks (e.g., $\tau$-bench, SST-2, SST-5), DLM-optimized prompts consistently improve the performance of a frozen target LLM (e.g., GPT-4o-mini). We further show that moderate diffusion step counts provide the best balance between refinement quality and stability. These results highlight diffusion-based prompt optimization as a general, model-agnostic, and scalable approach for enhancing LLM performance through iterative prompt refinement.\\
\begin{center}
\begin{tabular}{rll}
    \github & \textbf{\small{Code}} & 
    \href{https://github.com/SalesforceAIResearch/enterprise-deep-research}{\small{\texttt{https://github.com/SalesforceAIResearch/DLMForPromptOPT}}}
    \\
\end{tabular}
\end{center}
\end{tcolorbox}

\color{black}

\section{Introduction}
Large language models (LLMs) have demonstrated remarkable capabilities across a broad range of natural language understanding and generation tasks~\citep{naveed2025comprehensive,zhao2023survey,achiam2023gpt}. However, their performance is often highly sensitive to the quality of prompts used to elicit desired behaviors~\citep{liu2023pre,bai2024beyond}. This challenge has motivated extensive research on prompt optimization, ranging from manual prompt engineering~\citep{giray2023prompt} to automated approaches~\citep{shin2020autoprompt,yuksekgonul2024textgrad} that learn discrete or continuous prompts. While these methods have shown strong empirical results, they remain largely constrained to autoregressive models, which generate text sequentially and lack an efficient mechanism to revise or adapt prompts dynamically based on model outputs and user feedback.

\begin{figure}[htbp]
    \centering
    \includegraphics[width=\linewidth]{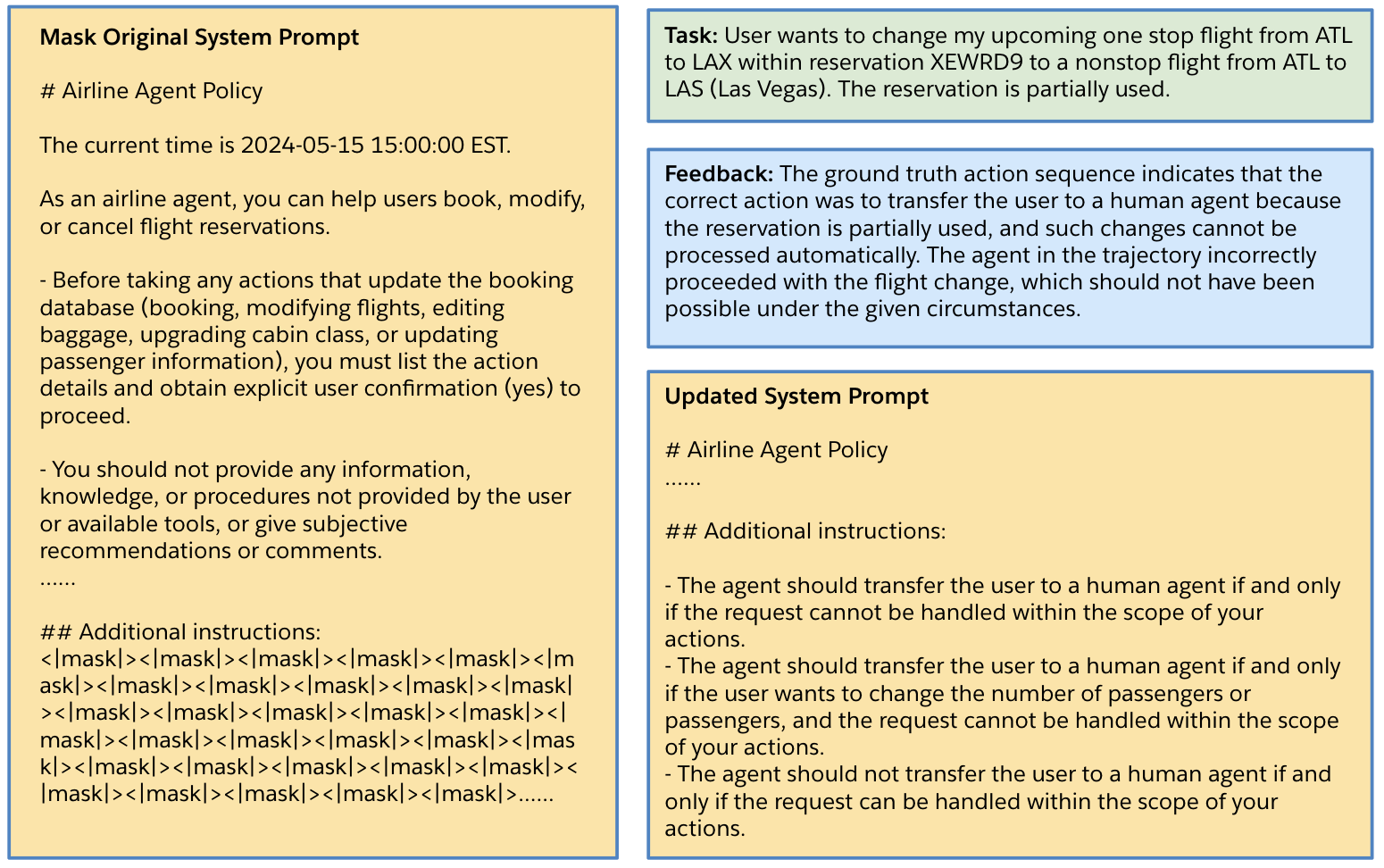}
    \caption{An example that illustrates prompt optimization by Diffusion Language Models (DLMs). In this case, we added additional instructions to original system prompt of $\tau$-bench airline. We leveraged DLMs to iteratively update the system prompt conditional on model output and the feedback by LLMs evaluator until all mask tokens are unmasked and predicted.}
    \label{fig:example}
\end{figure}

Diffusion language models (DLMs) offer a promising alternative~\citep{nie2025large,nie2024scaling,ye2025dream,chen2025coda,arriola2025block}. Unlike autoregressive LMs, which commit to tokens in a left-to-right manner, DLMs generate text through iterative refinement, allowing selective masking and regeneration of subsequences conditioned on context. This property makes DLMs naturally suited for interactive prompt optimization. In particular, given an interaction trace of the form [system prompt, user query, model output, feedback], a DLM can directly mask the system prompt and iteratively unmask or refine it while conditioning on the user query, observed model output, and associated feedback. Here, feedback is optional and can be any user feedback or from LLMs evaluator based on model outputs. Such an approach enables the system prompt to be adaptively optimized in light of both user intent and model behavior, addressing limitations of existing prompt optimization techniques that lack this form of dynamic adaptability. An example has been illustrated in Figure~\ref{fig:example} via a task in $\tau$-bench~\citep{yao2024tau}, where we added additional instructions attached to original system prompt. At the beginning, we added 100 mask tokens (i.e., <|mask|>) under "$\#\#$ Additional instructions" in the system prompt. Then we prompt DLMs (i.e., Dream-7B~\citep{ye2025dream}) to iteratively refine the system prompt. In the end, more detailed instructions according to the feedback provided by GPT-4o are generated by DLMs so that the system prompt is updated.

In this work, we investigate the integration of prompt optimization with diffusion-based language models. Building on prior studies of parameter-efficient prompt tuning and recent advances in diffusion text generation, we propose a framework that leverages the mask-and-refine architecture of DLMs to optimize system prompts conditioned on user interactions. Our approach extends prompt optimization beyond static or offline methods, introducing a dynamic, feedback-driven mechanism for aligning prompts with user intent. By systematically studying this integration, we aim to demonstrate that diffusion models are not only viable generative architectures but also uniquely effective for iterative prompt refinement in interactive LLM systems.
\section{Related Works}
Prompt optimization has emerged as a key technique for maximizing the performance of large language models. Early approaches relied on manual prompt design – carefully crafted task descriptions or examples – to induce models like GPT-3 to perform new tasks through zero-shot or few-shot learning. To reduce reliance on trial-and-error design, researchers developed automated prompt optimization methods. One line of work focuses on discrete prompt engineering. For example, AutoPrompt~\citep{shin2020autoprompt} uses gradient-guided search to discover token triggers that elicit desired behaviors from pretrained models. Complementing this, a family of continuous prompt tuning techniques has been introduced, often called “p-tuning” methods, such as prefix tuning~\citep{li2021prefix} and prompt tuning~\citep{lester2021power}, which freeze the original model’s weights and learn a small set of soft prompt embeddings to steer the model. Such learned prompts can incorporate signal from labeled data via backpropagation, and their efficacy grows with model size.

Diffusion language models (DLMs)~\citep{nie2025large, ye2025dream, nie2024scaling} have recently been proposed as an alternative to autoregressive decoding, replacing left-to-right generation with an iterative refinement process. By repeatedly denoising an initially corrupted sequence, DLMs are able to consider global context during generation, leading to improved fluency, robustness, and controllability. Unlike autoregressive models that irrevocably commit to earlier tokens, DLMs can flexibly mask and regenerate subsequences, enabling revisions of specific components in response to conditioning signals. This property naturally extends to the setting of prompt optimization: given an interaction trace $[\text{system prompt, user query, model output, feedback}]$, a DLM can mask the system prompt while conditioning on the user query, observed output, and associated feedback, and subsequently unmask or refine the prompt to improve alignment with user intent. This capability positions DLMs as particularly well-suited for dynamic prompt optimization scenarios that require revisiting and adapting the system prompt in light of downstream behavior.

\section{Methods}
\subsection{Diffusion Language Models}
Diffusion Language Models (DLMs) are a class of language models defined by a forward noising process and a backward denoising process~\citep{ye2025dream}. Specifically, forward process $q(x_{1:T}|x_0)=\prod_{t=1}^Tq(x_t|x_{t-1})$ progressively corrupts the original token sequence $x_0$ into a series of increasingly noisy sequences $x_{1:T}=x_1, ..., x_T$. At each step, original tokens are replaced with mask tokens (i.e., <|mask|>), gradually destroying the semantic content of the sequence. The forward process ends when the sequence is fully masked. The backward process is trained to progressively denoise the corrupted sequence and recover samples from the original data distribution. Starting from step $T$ down to step $0$, the model iteratively predicts the masked tokens at each step given a predefined step size. This reverse process can be formulated as
\begin{align}
    p_\theta(x_0) = \sum_{x_{1:T}\sim q} p(x_T)\prod_{t=1}^Tp_\theta(x_{t-1}|x_t)
\end{align},

where $\theta$ denotes the learnable model parameters optimized during training. A key property of DLMs that makes them promising for prompt optimization is their ability to generate tokens bidirectionally. This allows us to optimize components such as the system prompt—typically located at the beginning of a chat—while conditioning on the rest of the conversation.

\subsection{Prompt Optimization by Diffusion Language Models}

\begin{figure}[htbp]
    \centering
    \includegraphics[width=0.4\linewidth]{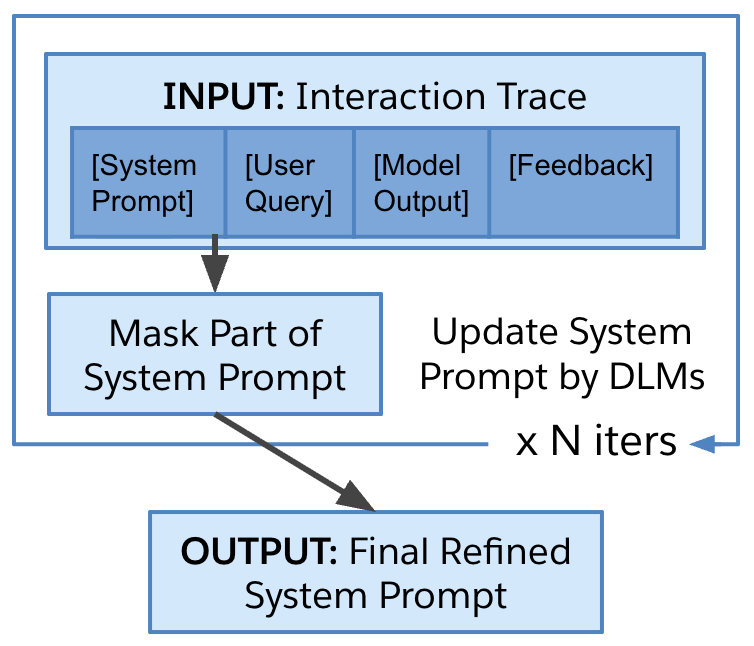}
    \caption{Overview of the iterative prompt optimization process using Diffusion Language Models (DLMs). The model iteratively masks and refines parts of the system prompt conditioned on the interaction trace. Feedback in the trace is optional and can usually be obtained either by user or by prompting LLMs-based evaluators.}
    \label{fig:po_dlm}
\end{figure}

We leverage DLMs for prompt optimization by directly updating the system prompt. A standard chat sequence can be decomposed as $x_0 = [x_{system}|x_{user}|x_{model}]$ where each component corresponds to the system prompt, user query, and model response, respectively. Optionally, additional feedback $x_{feedback}$ may be provided either by the user or by an LLM-based evaluator, yielding $x_0=[x_{system}|x_{user}|x_{model}|x_{feedback}]$. As illustrated in Figure~\ref{fig:po_dlm}, rather than randomly masking tokens in $x_0$ during the forward process, we selectively mask only the portion of $x_{system}$ targeted for optimization. This can be done by masking specific tokens or by inserting additional mask tokens to create editable positions.

As illustrated in Figure~\ref{fig:po_dlm}, the system prompt can be refined iteratively. At each iteration, a portion of the system prompt is masked and mask tokens are added, and the DLM is used to optimize the masked spans through the denoising process. This procedure is repeated until a predefined number of iterations $N$ is reached.
\section{Results}
We evaluate our approach on a diverse set of benchmarks spanning tool use, sentiment analysis, semantic similarity, and natural language inference. For function-calling tasks, we use $\tau$-bench, including the \textbf{Tau-bench-airline} and \textbf{Tau-bench-retail} subsets, which simulate realistic task-oriented dialogues requiring accurate structured function calls. For sentiment understanding, we adopt the Stanford Sentiment Treebank benchmarks (\textbf{SST-2} for binary classification and \textbf{SST-5} for fine-grained sentiment). We assess semantic equivalence using the \textbf{Microsoft Research Paraphrase Corpus (MRPC)} and evaluate reasoning over sentence relationships with the \textbf{Stanford Natural Language Inference (SNLI)} dataset, which requires predicting entailment, contradiction, or neutrality between sentence pairs.

We evaluate prompt optimization with DLMs (i.e., Dream-7B in experiments) from two perspectives. First, we compare their performance against traditional approaches, including TextGrad and prompting autoregressive models of comparable size (e.g., Llama 3-8B, Qwen3-8B). Secondly, we analyze the effectiveness of prompt optimization across varying numbers of diffusion steps.

\begin{table}[ht]
\centering
\caption{Success rate across domains. We use text diffusion models (i.e., Dream-7B), AR models (i.e., Qwen3-8B, Llama3-8B), and TextGrad to optimize prompt of each task. We compare success rate of GPT-4o-mini on prompt-optimized data and original data (i.e., Baseline).}
\begin{tabular}{lcccccc}
\toprule
\textbf{Model} & \textbf{Tau-bench-airline} & \textbf{Tau-bench-retail} & \textbf{SST2} & \textbf{SST5} & \textbf{MRPC} & \textbf{SNLI}\\
\midrule
Dream-7B   & 0.50 & 0.46 & 0.97 & 0.67 & 0.69 & 0.93\\
Llama3-8B & 0.41 & 0.42 & 0.96 & 0.63 & 0.69 & 0.92 \\
Qwen3-8B   & 0.42 & 0.46 & 0.96 & 0.65 & 0.69 & 0.92 \\
TextGrad   & 0.50 & 0.45 & 0.97 & 0.67 & 0.70 & 0.93 \\\hline
Baseline   & 0.43 & 0.42 & 0.93 & 0.55 & 0.61 & 0.88\\
\bottomrule
\label{tab:quant}
\end{tabular}
\end{table}

\paragraph{Quantitative Evaluation.}
Following Figure~\ref{fig:po_dlm}, we use Dream-7B~\citep{ye2025dream} as the DLM to optimize system prompts for each task. We then evaluate GPT-4o-mini using the optimized prompts and compare its performance against the original prompts (Baseline). As shown in Table~\ref{tab:quant}, prompt optimization consistently improves performance across all domains. Notably, Dream-7B achieves substantial gains on more challenging reasoning and structured-generation tasks. On Tau-bench-airline and Tau-bench-retail, success rates improve from 0.43/0.42 to 0.50/0.46, indicating that DLM-based prompt refinement enhances the model’s ability to produce accurate function calls in task-oriented settings. Similar trends are observed on natural language understanding benchmarks: performance on SST-5 increases markedly from 0.55 to 0.67, suggesting improved sensitivity to fine-grained sentiment distinctions, while gains on MRPC (0.61 $\rightarrow$ 0.69) and SNLI (0.88 $\rightarrow$ 0.93) demonstrate enhanced semantic matching and inference capabilities.

We further compare DLM-based optimization with AR prompt optimizers (Llama3-8B and Qwen3-8B) and gradient-based prompt editing (TextGrad). They all outperform the baseline with consistent improvements across diverse tasks, particularly on structured tool-use and fine-grained sentiment analysis. AR-based optimizers show competitive but slightly weaker gains, possibly due to their left-to-right generation constraint limiting flexible span-level edits. TextGrad performs comparably to Dream-7B on several tasks but requires gradient access and task-specific supervision, whereas DLM optimization operates purely through masked denoising. These results suggest that diffusion-based prompt optimization provides a robust and generalizable mechanism for improving downstream model performance without modifying the target model’s parameters.

\begin{figure}[htbp]
    \centering
    \includegraphics[width=0.8\linewidth]{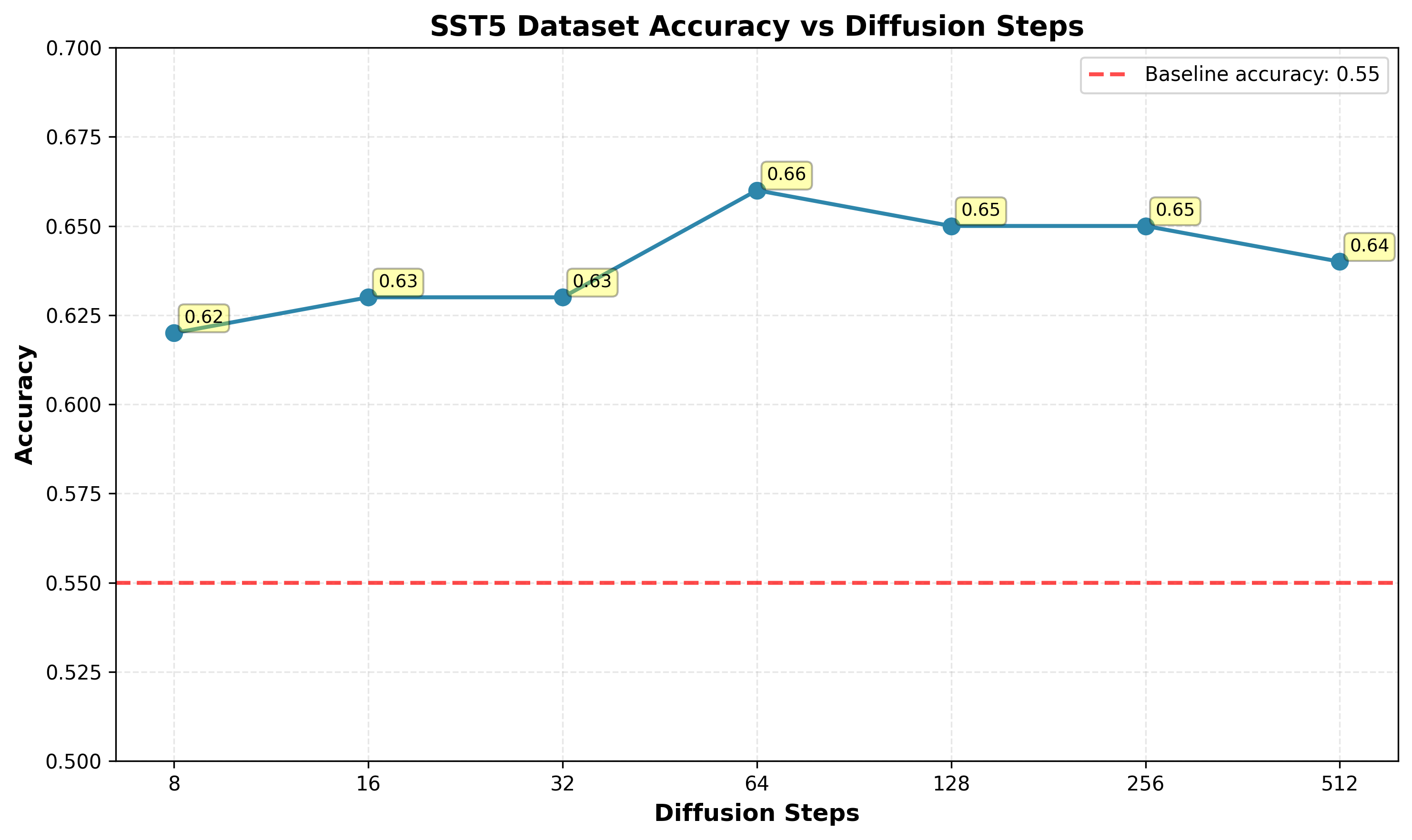}
    \caption{Accuracy on SST-5 of GPT-4o-mini vs. number of diffusion steps. The red dashed line is the performance of baseline.}
    \label{fig:sst5_steps}
\end{figure}

\paragraph{Effect of Diffusion Steps.}
Figure~\ref{fig:sst5_steps} illustrates the impact of the number of diffusion steps used by the DLM (i.e., Dream-7B) during prompt optimization on SST5 accuracy. We observe a clear improvement over the baseline accuracy of 0.55 (red dashed line) across all tested step counts, demonstrating that denoising iterations can yield meaningful gains. Performance increases steadily from 0.62 at 8 steps to a peak of 0.66 at 64 steps, suggesting that additional refinement iterations initially help the DLM produce higher-quality prompt updates. Beyond this point, however, gains plateau and slightly decline (e.g., 0.65 at 128–256 steps and 0.64 at 512 steps), indicating diminishing returns from excessive diffusion steps. This trend suggests a trade-off between refinement depth and over-editing. Moderate diffusion lengths provide sufficient opportunity for targeted prompt improvement, while too many steps may introduce unnecessary perturbations or drift. For example, we observed repeated contents generated by the DLM when diffusion steps go beyond 128. Overall, the results highlight that prompt optimization with DLMs is robust to a wide range of step counts, with around 64 steps offering a favorable balance between performance and computational cost in the case of SST5.

\section{Conclusion}

In this work, we introduced a diffusion-based framework for prompt optimization, where system prompts are iteratively refined through masked denoising using DLMs. Unlike AR or gradient-based prompt editing methods, our approach enables flexible span-level modifications without requiring access to model gradients or parameter updates of the target LLM. Experiments across function-calling, sentiment analysis, semantic similarity, and natural language inference benchmarks demonstrate that DLM-optimized prompts consistently improve the performance of a frozen downstream model (e.g., GPT-4o-mini). We further show that prompt quality benefits from iterative refinement with a moderate number of diffusion steps, highlighting an effective balance between optimization depth and stability. Overall, our results suggest that diffusion-based prompt optimization is a general, model-agnostic, and scalable strategy for enhancing LLM performance through prompt refinement alone.

\bibliographystyle{abbrvnat}
\bibliography{main}

\end{document}